\documentclass[conference]{IEEEtran}
\usepackage{graphicx}
\usepackage[style=ieee,backend=biber,sorting=none]{biblatex}
\usepackage{hyperref}
\usepackage{placeins}
\usepackage{balance}
\begin{document}
\title{Enhancing Understanding Through Wildlife Re-Identification} 
\author{J. Buitenhuis}
\maketitle              
\begin{abstract}
We explore the field of wildlife re-identification by implementing an MLP from scratch using NumPy, A DCNN using Keras, and a binary classifier with LightGBM for the purpose of learning for an assignment. Analyzing the performance of multiple models on multiple datasets. We attempt to replicate prior research in metric learning for wildlife re-identification. Firstly, we find that the usage of MLPs trained for classification, then removing the output layer and using the second last layer as an embedding was not a successful strategy for similar learning; it seems like losses designed for embeddings such as triplet loss are required. The DCNNS performed well on some datasets but poorly on others, which did not align with findings in previous literature. The LightGBM classifier overfitted too heavily and was not significantly better than a constant model when trained and evaluated on all pairs using accuracy as a metric. The technical implementations used seem to match standards according to comparisons with documentation examples and good results on certain datasets. However, there is still more to explore in regards to being able to fully recreate past literature.

\end{abstract}

\section{Introduction}

Wildlife re-identification is necessary for monitoring various facets of their well-being, used in behavioural studies and wildlife management 
\autocite{Schofield_Papafitsoros_Chapman_Shah_Westover_Dickson_Katselidis_2022,Vidal_2021}. Many techniques have been employed while tackling this problem, such as RFID \autocite{Bonter_Bridge_2011}, capturing and recapturing \autocite{royle2013spatial}, GPS tracking \autocite{BAUDOUIN201536}. These approaches all accomplish the goal but to varying degrees of cost and harm to the animals. Research shows that some species respond very differently and negatively to being tagged \autocite{birdstag}. Image-based techniques are the least intrusive but naturally have a poorer showing in terms of accuracy. Computer vision for the purpose of animal identification has been used as early as the 1990s and tends to follow a methodology of feature extraction (sometimes handcrafted as opposed to deep features which emerge as a byproduct as the network learns)\autocite{Schneider_Taylor_Linquist_Kremer_2019, DeepLearningCameraaTrap2018}.
\newline\newline
We should also establish the goal of the paper is to explore similarity learning via deep metric learning in wildlife re-identification instead of just classification. When models are trained for just classification, they have a fixed number of classes they can assign probability towards, meaning when you need to add and classify new classes, you need to start training again. When trying to learn metrics for similarity learning, a sufficiently well-performing model does not need to be retrained when new classes are introduced or old classes removed \autocite{meyer2019importance}.
\newline
The field continues to improve its results by finding better loss functions and better neural network architectures \autocite{Datasets}. However, there is no single network that always outperforms the others, at least when compared using modern metrics for deep metric learning \autocite{zylNkosi}. However, When using traditional classification accuracy, some models do seem to be measurably better no matter the dataset \autocite{Cermak_2024_WACV}.
\newline
One aspect that seems underrepresented in wildlife re-identification but useful in similar tasks, such as animal species classification and animal detection, is the use of synthetic training data. In species classification, research shows that synthetic data can improve accuracy on rare classes \autocite{beery2019synthetic}. Synthetic data also shows promising results in animal detection \autocite{Bonetto_2023}. 
\newline\newline
Recently, major improvements have been made in diffusion-based model image generation, especially in creating consistent characters, such as dreambooth \autocite{ruiz2022dreambooth} and the use of low-rank adaptation to customize image generation models\autocite{LoraForDiffusion}. It seems as though these models have the potential to generate synthetic data that can be used to improve the extraction of features. 
\begin{figure}
    \centering
    \includegraphics[width=0.8\linewidth]{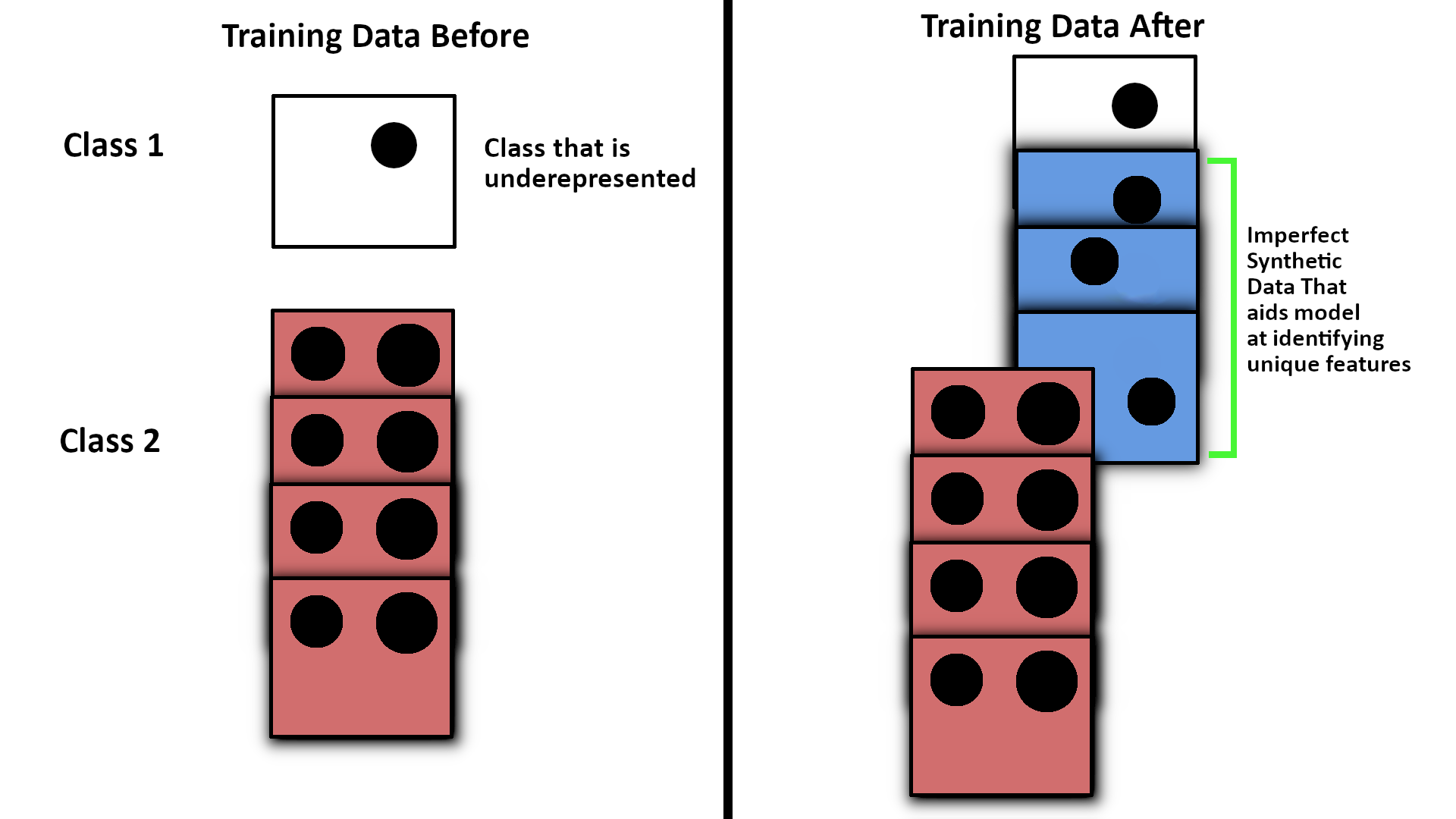}
    \caption{Example representing the idea behind introducing synthetic training data}
    \label{fig:enter-label}
\end{figure}

I think one reason why this is less explored is that the generative models themselves rely on some form of feature extraction such as CLIP \autocite{schuhmann2021laion}, so using these generations ends up being a sort of transfer learning, which, of course, can be done efficiently by just using pre-trained backbones for models. 

\section{Experimental Design}
A number of techniques have been developed to improve wildlife re-identification results. We conduct a benchmark of several models across multiple datasets using a Pytorch library \autocite{Musgrave2020PyTorchML} that can calculate MAP@R matching previously established methods \autocite{zylNkosi}. 

 \subsection{Datasets}
 All data can be found from the recently published tool \autocite{Datasets}. In particular, we make use of the lion data originally shared in 2021 by Dlamini et al. \autocite{zylNkosi}. 
 \subsection{Experimental Setup}
    \subsubsection{MLP from scratch with NumPy}
    The MLPs used are not trained for metric learning directly but for classification, utilizing a single hidden layer with a RELU activation function followed by a softmax layer into the appropriate number of classes. In the case of LionData, which was organized more for the purpose of metric learning, we have 75 classes in the training data. In the case of FriesanCattle2017, we separate nine classes out of the 89 for testing. Hyper-parameters for the number of batches and learning rate are automatically selected for all MLPS using a simple grid search for the highest 5-fold cross-validation mean accuracy. The model is then trained with early stopping with a patience of 5 and evaluated with 5-fold cross-validation mean accuracy. All our MLPs use our implementation of the ADAM optimizer with $\beta_1=0.9 \ \beta_2=0.999 \ \epsilon=1e-8$. Tests are conducted with the same fixed seed for all methods. Input images are converted to greyscale and scaled to 64x64. 
    \subsubsection{Keras/TF DCNNs}
    The TensorFlow CNNs are all trained using triplet loss with a batch size of 32 for 3 epochs with 100 steps per epoch. The inputs are converted to floating points between -1 and 1 during training and testing. 
    \subsubsection{Wide models}
    We trained wide models for identifying whether two embeddings of classes should be considered different or not, essentially making them distance functions; the features are computed using OpenCV, namely a combination of ORB, SIFT, and BRISK. The lightGBM model is trained once the default hyper-parameters are used. The model is trained on every pair of training data with a matching label indicating if the images are the same or not. The tests are also done with every pair of test data. 
    \subsubsection{Recording metric learning metrics}
    We utilize a PyTorch library for computing metrics \autocite{Musgrave2020PyTorchML}; regardless of the model or data used, we pass all embeddings to the same metric measuring function. MAP@R for LionData was averaged over 14 runs. \\
    Other modifications/variations will be indicated in the title of the model, such as when we employed two hidden layers or basic data augmentation. 
 \subsection{Analysis}
    With motivation from Musgrave et al. \autocite{musgrave2020metric}, we compute MAP@R. In particular, we use an average accuracy per class as described in Musgrave's \href{https://kevinmusgrave.github.io/pytorch-metric-learning/accuracy_calculation/}{documentation} stating
    , "This can be useful if your dataset has unbalanced classes" which is certainly the case for FriesianCattle2017 where some classes only have 1 example, and others have up to 47. 
 \subsection{Tools and Libraries}
Pytorch, Pytorch Libraries \autocite{Musgrave2020PyTorchML}, wildlife-datasets loader \autocite{Datasets}. TensorFlow Similarity \autocite{TFSIM}. \href{https://github.com/numpy/numpy}{Numpy}. 

\newpage
\section{Experimental Results}
\subsection{MLP from scratch}
\FloatBarrier
\begin{table}[!h]
\caption{Metrics for various MLPs on FriesianCattle2017}\label{tab2}
\centering
\begin{tabular}{l|l|l}
    \hline Method & MAP@R\% & Accuracy \\
    \hline 64x2 neuron hidden layer \ & 22.28 & 48.58 \\
    \hline 64 neuron hidden layer \ & 26.74 & 41.25\\
    \hline 32 neuron hidden layer \ & 27.53 & 47.97\\ 
    \hline 16 neuron hidden layer \ & 21.22 & 23.79\\ 
    \hline
\end{tabular}
\end{table}

These are not incredible results for accuracy, but they are certainly useful, considering there are 80 classes. In terms of MAP@R, these values may seem respectable. However, the FriesanCattle2017 data has such unique images for each class that, in fact, a completely randomly instantiated MLP with a 64 embedding Relu layer was able to obtain a MAP@R of 29\%. 

\subsection{MAP@R\% for various DCNNS In TF}
\begin{table}[!h]
\caption{Results for various CNN's on FriesanCattle2017}\label{tab3}
\centering
\begin{tabular}{l|l}
    \hline Method & MAP@R\% \\
    \hline ResNet50V2 Backbone 128 DIM \ & 49.90 \\
    \hline ResNet50V2 Backbone 64 DIM \ & 47.78 \\
    \hline ResNet50V2 Backbone 32 DIM \ & 47.57 \\
    \hline ResNet50V2 Backbone 16 DIM \ & 45.39 \\
    \hline
\end{tabular}
\end{table}

Seemingly, the data is so clean that despite there sometimes being only 1 example per class, the model is able to create effective embeddings with only 16 dimensions. 
\FloatBarrier
\begin{table}[!h]
\caption{Results for various CNN's on LionData}\label{tab4}
\centering
\begin{tabular}{l|l}
    \hline Method & MAP@R\% \\
    \hline ResNet50V2 Backbone 32 DIM \ & 0.87 \\
    \hline ResNet50V2 Backbone 64 DIM (2 runs) (Augmentation) \ & 0.74 \\
    \hline ResNet50V2 Backbone 64 DIM (2 runs) \ & 0.93 \\
    \hline
\end{tabular}
\end{table}
Surprisingly, the results were poor in contrast to the FriesanCattle2017 data. The augmentation was applied in the form of preprocessing images by random flipping and random rotation. 
\subsection{Wide models}
For FriesanCattle2017, we train a lightGBM binary classifier that takes two embeddings of hardcoded features and decides whether they are the same animal or not; this model is able to obtain 80.82\% on the test examples. However, this result is not significantly better than a model that simply guesses that every pair is different since, naturally, when testing every pair, most pairs will not be from the same class. 

\section{Discussion}
It seems that the approach of training an MLP for classification, removing the last classification layer, and using the last hidden layer as an embedding was not successful despite the models succeeding in learning classification even with a very small capacity. Indicating that perhaps more capacity is needed for meaningful representations to appear in the hidden layers. This is a shortcoming of this research where, regrettably, we could not explore bigger models due to time and computational constraints. 
\\
The DCNNS, as expected, perform significantly better than the MLPs. On the FriesanCattle2017 data, the MAP@R values are generally good when compared to other SOTA MAP@R values on different datasets \autocite{musgrave2020metric,zylNkosi}.  However, it displays strikingly poor results regarding LionData MAP@R when compared to Dlamini et al. \autocite{zylNkosi} where in table 4 the best MAP@R\% obtained with DenseNet-201 backbone was 20.8 and a similar ResNet-152 backbone scores 17.3 when trained on Triplet loss with semi-hard mining just as our models are however the training approaches are not identical, we employ a soft margin for Triplet loss as opposed to a fixed margin of 0.2 however we believe the use of a soft margin is well motivated \autocite{hermans2017defense}. However, we employ far fewer epochs, and whilst three epochs were enough for FriesianCattle2017, it might not be enough for more complex data with limited examples from varying perspectives like LionData. I do not believe our implementation of MAP@R is faulty since when compared with examples from the Pytorch-Metric-Learning Library our MAP@R function that works with Numpy arrays differs only slightly from the Pytorch tensor examples specifically the example obtains 0.91715916290391 whilst we obtain 0.9172336161818239 we suspect the difference is due to floating point inaccuracies when transferring the CUDA tensors to Numpy arrays where the distances are computed on the CPU instead. 
\\
Regrettably, we would have liked to explore more; however, due to time/computational limits, we fell short of exploring all aspects.

\section{Conclusion}
The goal of this research was to explore ideas in metric learning and the Wildlife re-identification field at large by implementing two baseline solutions done from scratch and two off-the-shelf SOTA solutions. The MLP with cross-validation and early stopping coupled with a hyper-parameter grid search was able to easily learn classification, but the second to last layer did not contain information that was meaningful for metric learning. The choice of DCNNs was to attempt to replicate previous literature, which was not successful despite the models performing well on other datasets; we could not replicate \autocite{zylNkosi}. We confirmed our implementation of MAP\@R closely matches documentation, so more research is needed to figure out why the DCNN performed so poorly on the Lion data. \newline The Wide models heavily over-fitted the training data, not performing significantly better than constant models, suggesting a better training methodology and/or a better feature selection. \newline This study is limited by the fact that it is designed for understanding these models within the context of a semester course and, as such, is limited by model selections and time/computation availability. In the future, I'd like to investigate transformer models for wildlife identification since these models have recently been shown to compete well on classification-based tasks and might perform even better on similarity learning tasks with metrics such as MAP@R \autocite{Datasets}. 
\balance
\printbibliography

@article{birdstag,
author = {Carley R. Schacter and Ian L. Jones},
title = {{Effects Of Geolocation Tracking Devices On Behavior, Reproductive Success, and Return Rate of Aethia Auklets: An Evaluation of Tag Mass Guidelines}},
volume = {129},
journal = {The Wilson Journal of Ornithology},
number = {3},
publisher = {The Wilson Ornithological Society},
pages = {459 -- 468},
year = {2017},
doi = {10.1676/16-084.1},
URL = {https://doi.org/10.1676/16-084.1}
}

@InProceedings{Cermak_2024_WACV,
    author    = {\v{C}erm\'ak, Vojt\v{e}ch and Picek, Lukas and Adam, Luk\'a\v{s} and Papafitsoros, Kostas},
    title     = {WildlifeDatasets: An Open-Source Toolkit for Animal Re-Identification},
    booktitle = {Proceedings of the IEEE/CVF Winter Conference on Applications of Computer Vision (WACV)},
    month     = {January},
    year      = {2024},
    pages     = {5953-5963}
}

@article{Schofield_Papafitsoros_Chapman_Shah_Westover_Dickson_Katselidis_2022, title={More aggressive sea turtles win fights over foraging resources independent of body size and years of presence}, volume={190}, DOI={10.1016/j.anbehav.2022.05.006}, journal={Animal Behaviour}, publisher={Elsevier BV}, author={Schofield, Gail and Papafitsoros, Kostas and Chapman, Chloe and Shah, Akanksha and Westover, Lucy and Dickson, Liam C.D. and Katselidis, Kostas A.}, year={2022}, month=aug, pages={209–219} }

@article{Vidal_2021,
   title={Perspectives on Individual Animal Identification from Biology and Computer Vision},
   volume={61},
   ISSN={1557-7023},
   url={http://dx.doi.org/10.1093/icb/icab107},
   DOI={10.1093/icb/icab107},
   number={3},
   journal={Integrative and Comparative Biology},
   publisher={Oxford University Press (OUP)},
   author={Vidal, Maxime and Wolf, Nathan and Rosenberg, Beth and Harris, Bradley P and Mathis, Alexander},
   year={2021},
   month=may, pages={900–916} }

@article{Bonter_Bridge_2011, title={Applications of radio frequency identification (RFID) in ornithological research: a review}, volume={82}, DOI={10.1111/j.1557-9263.2010.00302.x}, number={1}, journal={Journal of Field Ornithology}, publisher={Resilience Alliance, Inc.}, author={Bonter, David N. and Bridge, Eli S.}, year={2011}, month=feb, pages={1–10} }

@article{BAUDOUIN201536,
title = {Identification of key marine areas for conservation based on satellite tracking of post-nesting migrating green turtles (Chelonia mydas)},
journal = {Biological Conservation},
volume = {184},
pages = {36-41},
year = {2015},
issn = {0006-3207},
doi = {https://doi.org/10.1016/j.biocon.2014.12.021},
url = {https://www.sciencedirect.com/science/article/pii/S000632071400500X},
author = {Marie Baudouin and Benoît {de Thoisy} and Philippine Chambault and Rachel Berzins and Mathieu Entraygues and Laurent Kelle and Avasania Turny and Yvon {Le Maho} and Damien Chevallier}
}

@book{royle2013spatial,
  title={Spatial Capture-Recapture},
  author={Royle, J.A. and Chandler, R.B. and Sollmann, R. and Gardner, B.},
  isbn={9780124071520},
  lccn={2013030684},
  url={https://books.google.co.za/books?id=RO08-S-amZMC},
  year={2013},
  publisher={Elsevier Science}
}

@misc{beery2019synthetic,
      title={Synthetic Examples Improve Generalization for Rare Classes}, 
      author={Sara Beery and Yang Liu and Dan Morris and Jim Piavis and Ashish Kapoor and Markus Meister and Neel Joshi and Pietro Perona},
      year={2019},
      eprint={1904.05916},
      archivePrefix={arXiv},
      primaryClass={cs.CV}
}

@article{Schneider_Taylor_Linquist_Kremer_2019, title={Past, present and future approaches using computer vision for animal re‐identification from camera trap data}, volume={10}, DOI={10.1111/2041-210x.13133}, number={4}, journal={Methods in Ecology and Evolution}, publisher={Wiley}, author={Schneider, Stefan and Taylor, Graham W. and Linquist, Stefan and Kremer, Stefan C.}, year={2019}, month=jan., pages={461–470} }

@inproceedings{Bonetto_2023,
   title={Synthetic Data-Based Detection of Zebras in Drone Imagery},
   url={http://dx.doi.org/10.1109/ECMR59166.2023.10256293},
   DOI={10.1109/ecmr59166.2023.10256293},
   booktitle={2023 European Conference on Mobile Robots (ECMR)},
   publisher={IEEE},
   author={Bonetto, Elia and Ahmad, Aamir},
   year={2023},
   month=sep }

@article{DeepLearningCameraaTrap2018, title={Automatically identifying, counting, and describing wild animals in camera-trap images with deep learning}, volume={115}, DOI={10.1073/pnas.1719367115}, number={25}, journal={Proceedings of the National Academy of Sciences}, publisher={Proceedings of the National Academy of Sciences}, author={Norouzzadeh, Mohammad Sadegh and Nguyen, Anh and Kosmala, Margaret and Swanson, Alexandra and Palmer, Meredith S. and Packer, Craig and Clune, Jeff}, year={2018}, month=jun. }

@Article{zylNkosi,
AUTHOR = {Dlamini, Nkosikhona and van Zyl, Terence L.},
TITLE = {Comparing Class-Aware and Pairwise Loss Functions for Deep Metric Learning in Wildlife Re-Identification},
JOURNAL = {Sensors},
VOLUME = {21},
YEAR = {2021},
NUMBER = {18},
ARTICLE-NUMBER = {6109},
URL = {https://www.mdpi.com/1424-8220/21/18/6109},
PubMedID = {34577319},
ISSN = {1424-8220},
DOI = {10.3390/s21186109}
}

@article{ruiz2022dreambooth,
  title={DreamBooth: Fine Tuning Text-to-image Diffusion Models for Subject-Driven Generation},
  author={Ruiz, Nataniel and Li, Yuanzhen and Jampani, Varun and Pritch, Yael and Rubinstein, Michael and Aberman, Kfir},
  booktitle={arXiv preprint arxiv:2208.12242},
  year={2022}
}

@inproceedings{LoraForDiffusion,
 author = {Gu, Yuchao and Wang, Xintao and Wu, Jay Zhangjie and Shi, Yujun and Chen, Yunpeng and Fan, Zihan and XIAO, WUYOU and Zhao, Rui and Chang, Shuning and Wu, Weijia and Ge, Yixiao and Shan, Ying and Shou, Mike Zheng},
 booktitle = {Advances in Neural Information Processing Systems},
 editor = {A. Oh and T. Neumann and A. Globerson and K. Saenko and M. Hardt and S. Levine},
 pages = {15890--15902},
 publisher = {Curran Associates, Inc.},
 title = {Mix-of-Show: Decentralized Low-Rank Adaptation for Multi-Concept Customization of Diffusion Models},
 url = {https://proceedings.neurips.cc/paper_files/paper/2023/file/3340ee1e4a8bad8d32c35721712b4d0a-Paper-Conference.pdf},
 volume = {36},
 year = {2023}
}

@article{Musgrave2020PyTorchML,
  title={PyTorch Metric Learning},
  author={Kevin Musgrave and Serge J. Belongie and Ser-Nam Lim},
  journal={ArXiv},
  year={2020},
  volume={abs/2008.09164}
}

@InProceedings{Datasets,
    author    = {\v{C}erm\'ak, Vojt\v{e}ch and Picek, Luk\'a\v{s} and Adam, Luk\'a\v{s} and Papafitsoros, Kostas},
    title     = {{WildlifeDatasets: An Open-Source Toolkit for Animal Re-Identification}},
    booktitle = {Proceedings of the IEEE/CVF Winter Conference on Applications of Computer Vision (WACV)},
    month     = {January},
    year      = {2024},
    pages     = {5953-5963}
}

@article{TFSIM,
  title={TensorFlow Similarity: A Usable, High-Performance Metric Learning Library},
  author={Elie Bursztein and James Long and Shun Lin and Owen Vallis and Francois Chollet},
  journal={Fixme},
  year={2021}
}

@inproceedings{musgrave2020metric,
  title={A metric learning reality check},
  author={Musgrave, Kevin and Belongie, Serge and Lim, Ser-Nam},
  booktitle={Computer Vision--ECCV 2020: 16th European Conference, Glasgow, UK, August 23--28, 2020, Proceedings, Part XXV 16},
  pages={681--699},
  year={2020},
  organization={Springer}
}

@article{hermans2017defense,
  title={In defense of the triplet loss for person re-identification},
  author={Hermans, Alexander and Beyer, Lucas and Leibe, Bastian},
  journal={arXiv preprint arXiv:1703.07737},
  year={2017}
}

@misc{meyer2019importance,
      title={The Importance of Metric Learning for Robotic Vision: Open Set Recognition and Active Learning}, 
      author={Benjamin J. Meyer and Tom Drummond},
      year={2019},
      eprint={1902.10363},
      archivePrefix={arXiv},
      primaryClass={cs.CV}
}

@article{schuhmann2021laion,
  title={Laion-400m: Open dataset of clip-filtered 400 million image-text pairs},
  author={Schuhmann, Christoph and Vencu, Richard and Beaumont, Romain and Kaczmarczyk, Robert and Mullis, Clayton and Katta, Aarush and Coombes, Theo and Jitsev, Jenia and Komatsuzaki, Aran},
  journal={arXiv preprint arXiv:2111.02114},
  year={2021}
}

\end{document}